\documentclass[fleqn,10pt]{wlscirep}

\usepackage[utf8]{inputenc}
\usepackage[T1]{fontenc}
\usepackage{bm}
\usepackage{amsmath,amssymb}

\title{Personalized next-best action recommendation with multi-party interaction learning for automated decision-making}

\author[1,*]{Longbing Cao}
\author[1]{Chengzhang Zhu}
\affil[1]{Data Science Lab, University of Technology Sydney, Australia}

\affil[+]{these authors contributed equally to this work}

\begin{abstract}
Automated next-best action recommendation for each customer in a sequential, dynamic and interactive context has been widely needed in natural, social and business decision-making. Personalized next-best action recommendation must involve past, current and future customer demographics and circumstances (states) and behaviors, long-range sequential interactions between customers and decision-makers, multi-sequence interactions between states, behaviors and actions, and their reactions to their counterpart's actions. No existing modeling theories and tools, including Markovian decision processes, user and behavior modeling, deep sequential modeling, and personalized sequential recommendation, can quantify such complex decision-making on a personal level. We take a data-driven approach to learn the next-best actions for personalized decision-making by a reinforced coupled recurrent neural network (CRN). CRN represents multiple coupled dynamic sequences of a customer's historical and current states, responses to decision-makers' actions, decision rewards to actions, and learns long-term multi-sequence interactions between parties (customer and decision-maker). Next-best actions are then recommended on each customer at a time point to change their state for an optimal decision-making objective. Our study demonstrates the potential of personalized deep learning of multi-sequence interactions and automated dynamic intervention for personalized decision-making in complex systems.
\end{abstract}
\begin{document}

\flushbottom
\maketitle
%
%


\section*{Introduction}
In enterprise and complex problem-solving, automated and personalized decision-making is highly needed but rarely possible in practice. Personalized decision-making requires personalized next-best actions to be learned and used in a dynamic, sequential and interactive process and context, which is extremely demanding in both private and public sectors and natural and social systems. Examples are next-best treatments to-be-made by healthcare providers on patients, next-best trading strategies to-be-taken by investors in a capital market, next-best interventions on cybersecurity attacks or climate change in real time, next-best communications between a bank and its clients, and any other services involving client-provider interactions. Personalized next-best action-taking sets up a high standard for long-term dependent, dynamic, sequential and interactive personalized and automated decision-making in sophisticated and constrained real-life environments. However, automated decision-making with personalized next-best action recommendation is extremely challenging: (1) the circumstances and response behaviors of each target client (interchangeable with customer) must be characterized and modeled when they evolve over time; (2) any decision actions taken by a decision-maker on the client at a time step takes place in a sequential and interactive context, where both client responses and decision actions interact and co-evolve under client-specific circumstances and decision-policy constraints, forming multiple interactive and coupled sequences; (3) often multiple decision choices are available at a time step, and the best decision action needs to fit client states, expected decision goals and effect, and the underlying environment; (4) taking any decision actions will further affect the state, action and environment at the next time step, and the cumulative effect from all prior steps also evolves along the sequential action-response interactions, which form long-term dependencies between multiple sequences to affect the next-best action selection; and (5) while each next-best action is to achieve an expected local goal and effect on the client, the sequence of next-best actions should generate the optimal global goal and effect.

In practice, domain-driven action rules are often generated and tuned by a group of domain experts to address the complexities in the aforementioned personalized next-best action-taking in complex enterprises and systems. This domain-driven action selection collectively considers and balances the relationships between service policies, constraints, client circumstances, business procedures, risk indicators, decision rules, and intervention strategies. Although hand-crafted action rules may be effective for specific and static scenarios on a small scale, they are ad hoc and ineffective for wide and dynamic applications and for large-scale real-time decision-making. They lack a general and proactive capacity to tackle personalized, sequential and interactive decision-making and often result in issues such as a high false intervention rate, high missing rate, and low cost-effectiveness. 

The advances in new-generation artificial intelligence and data science have made possible the automated selection and optimal recommendation of personalized next-best actions in the above complex decision-making settings. This, however, poses a significant challenge to existing decision-support systems and modeling methods, including sequential decision-making \cite{puterman2014markov,korn2018heuristic}, sequential and personalized recommendation \cite{qian2014personalized,ji2015next,ChenY0YW20,WangX21}, and deep learning \cite{bengio2013representation,lecun2015deep}. To the best of our knowledge, there are no existing theories or modeling methods capable of handling the aforementioned demand and challenges in an automated or semi-automated manner. Typical sequential decision-making methods \cite{sutton1998reinforcement,bellman2013dynamic,chakraborty2014dynamic,mnih2015human,van2016deep} assume that decision-making falls in the Markov decision processes (MDPs), i.e., the next state only depends on the current state and action \cite{puterman2014markov}. Other approaches involve all historical states such as by weighing their impact on current states \cite{boutilier1995process,friedrich2011spatio,clarke2015human}. They do not fit personalized decision-making that goes beyond Markovian \cite{whitehead1995reinforcement,clarke2015human,PengZSHWT18}, which involves complex interactions and couplings between clients and providers and their states, responses and actions \cite{FagerbergJ95,IshigakiTSA18,ijcai_WangSC13,ThrunPU20,egorov2017pomdps,HanJTGZ18}, as well as their dynamics and adaptation to bi-party (or multi-party) interactions \cite{taghia2018uncovering,mcdonald2019bayesian}. More recent work selectively represents historical interactions between clients and decision-makers using methods such as temporal logic-based models \cite{bacchus1996rewarding,bacchus1997structured,thiebaux2006decision,brafman2017specifying,DuR0Z19}, and recurrent neural networks (RNNs) with memories \cite{hausknecht2015deep,bajor2017predicting}. However, they are ineffective for next-best action recommendation, since they either treat states and actions homogeneously, i.e., ignoring the differences between states and actions, or ignore their complex interactions and couplings, by taking a predefined action on a state without selecting the actions for the best fit between clients, states, actions, and contexts. In addition, personalized recommendation and sequential recommender systems (including next-item and next-basket recommendation) have emerged recently \cite{qian2014personalized,zuo2015personalized,WangS21,ji2015next,chou2016addressing} to recommend particular or next products to users who may prefer in the next context. The existing methods do not involve comprehensive user-product couplings and heterogeneities (i.e., non-IIDness of users and items \cite{Cao-rs-eng}), dynamic user-product interactions, sequential actions and responses, or optimal decision effects, etc. In addition, intensive research has been done on group decision making and recommendation \cite{MINER1984112,ShuJYW21,Zhang0JWZ17}, which are irrelevant to this work.

Here, we introduce a computational approach: a reinforced coupled recurrent network (CRN) to model the intrinsic nature of recommending personalized next-best actions in the aforementioned complex decision-making settings. CRN integrates deep learning, reinforcement learning, behavior informatics and recommender systems to learn dynamic, sequential, interactive and personalized decision-making processes. First, CRN models client circumstances, states, behaviors, responses and decision-making actions by multi-dimensional sequential representations using recurrent neural networks. This captures and transforms the states and behaviors of clients and actions made by decision-makers and their evolution into computable vector representations. Second, CRN builds a coupled recurrent unit (CRU) to capture relevant historical behaviors and simultaneously learn the following sophisticated couplings and interactions between clients and decision-makers on the above learned sequential representations using two long-term memories and five control gates: (1) the long-term sequential dependencies between an action and its previous actions taken by a decision-maker, called \textit{action-action dependence}, to reveal the influence and transition between a series of prior actions and the current action; (2) the long-term sequential dependencies between a response and its previous responses made by a client, called \textit{response-response dependence}, to learn the influence and transition between previous responses and the current one; and (3) the long-term sequential dependencies between a current response and its corresponding previous actions, called \textit{action-response dependence}, to model the influence and transition between previous sequential actions and the current response of a client. As a result, CRU captures, represents and memorizes a sequence of relevant interactions between a client and a decision-maker with their particular states and behaviors and their history. Third, CRN combines the represented behaviors with the client's current state features and transforms them to a compact client state representation, which models client states and their transition. Lastly, CRN models the reward to candidate actions and learns the dependence between the current reward to actions and the next client state in a compact state representation to determine the next-best action tailored for the client to achieve the decision goal. 

The CRN model was tested in a major Australian government agency for debt collection to recommend next-best intervention actions on specific debtors for tailored, active and efficient debt collection. CRN automatically recommends the next-best action tailored for each debtor at a particular time by incorporating the debtor's current state and historical records, the government's optional and constrained action sequences, and reward to actions specified by their debt collectors (domain experts) measuring the effectiveness of action on debt collection. In contrast to the related work that either assumes a Markovian property of sequential decision-making actions or has a limited computational capability in modeling complex contexts and interactions in personalized decision-making, our approach collectively involves and automatically learns sequences of decision actions, client behaviors and states, their interactions and transitions, the action-action, response-response and action-response dependencies, and the action effect (reward) on client responses in dynamic, sequential and interactive decision-making contexts at a client level.

\section*{Methods}
\label{sec:methods}

\subsection*{Learning next-best actions}
\begin{figure}[!h]
    \includegraphics[width=0.6\textwidth]{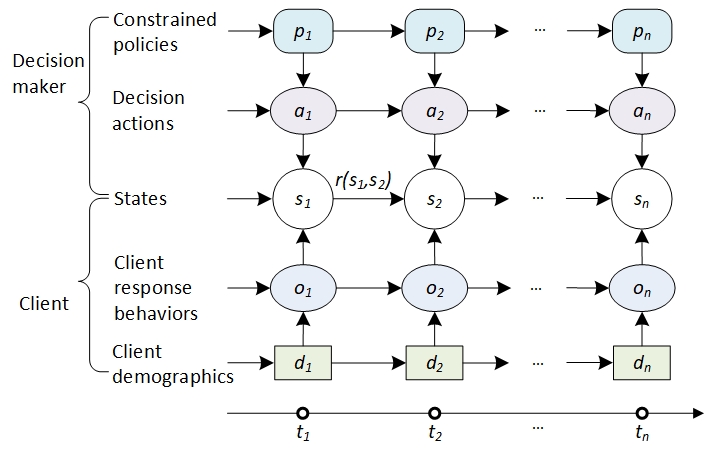}
    \caption{{\bf Next-best action-based personalized decision-making in constrained, tailored, sequential and interactive dynamic processes with state-action-response-coupled sequences.} A decision-maker interacts with a client sequentially at $t$ time steps. At each time point $i$, the client is associated with his demographics $d_i$ and state $s_i$. The decision-maker takes an action $a_i$ on the client's state $s_i$ under policy constraint $p_i$. The client responds to the action with his behavior $o_i$, and the undertaken action takes effect with reward $r_i$ showing the transform effect from the current state to the next state. Note, the diagram only illustrates an ideal scenario: one client who interacts with one decision-maker, and one action corresponds to one response at each time point. From bottom to top: a rectangle represents a client's demographics, a light-blue ellipse represents a client's response, a circle represents a client's states, a light-red ellipse represents decision actions, and a rounded rectangle represents policies constraining decision actions.}
    \label{fig:nbaexample}
\end{figure}

Assume a next-best action selection process (illustrated in Fig \ref{fig:nbaexample}) involves a client and their demographics and states, a decision-maker and their actions taken on the client under certain policy constraints, the response (behaviors) of the client to the actions, and the reward that measures the effectiveness of an action on the client to achieve business goals at a time point. For example, in  government services such as social welfare and taxation, when a client incurs a debt (called a debtor, i.e., a government client who owes money to the government), the government may take a series of actions to recover the debt in full or fast. Although debt collection is a widely used yet sophisticated process, experienced debt collectors not only consider a debtor's circumstances, the government's service policies and constraints, business objectives, and the effect of particular actions, they also monitor a debtor's responses to the implemented actions before a new action is taken. Some collectors may quantify the rewards for applied actions to indicate their effectiveness in intervention. At present, such action-based debt collection is mainly driven by business assumptions and rules, i.e., debt collection rules which we also call \textit{domain-driven action rules} for complex systems and decision-making \cite{aikp_Cao15}. Domain-driven action rules play an important role in active and personalized debt collection using the collectors' experience, understanding and belief of the debtors' circumstances and possible responses and judgment in matching actions with client profiles. 

However, domain-driven action selection is often ad hoc, costly and unsuitable for complex enterprise decision-making. A debt collection action must be carefully chosen and applied on a debtor at a particular time point by considering the client's circumstances, the government's policies and service objectives, the previous actions already taken on the client, the debtor's responses, the potential response to an action, and the business impact of interventions (e.g., whether the debt will be collected faster, in a less costly manner etc.). The action selection process also needs to consider a debtor's evolving circumstances, which further change during the sequential interactions with the government. Consequently, debt collection often involves a sequence of constrained candidate actions and the interactions with debtors in dynamic contexts sequentially and interactively. In summary, smart debt collection must be \textit{tailored} for each debtor and debt case, \textit{dynamic} in terms of catering for evolving debtor circumstances and business environmental settings (i.e., states), \textit{interactive} between debtors and debt collectors with their iterative communications over the collection process, and \textit{sequential} with both preceding and successive actions and states considered. 

We model the above debt collection problem illustrated in Fig \ref{fig:nbaexample} as \textit{personalized next-best action recommendation} on each client in a dynamic, sequential, interactive and constrained decision-making process. This personalized next-best action recommendation involves client information, sequences of client and decision-maker behaviors, and interactions between clients and the decision-maker under certain contexts and constraints at each time point.  

Without loss of generality, we assume a client $c$ over time $t$ can be described by a three-element tuple $C_t = <D_t, A_{t-1}, O_{t}>$, where $D_t = \{d_i| i=1, \cdots, n_d\}$ refers to a set of client's relatively stable information $d_i$ (e.g., the demographics of the client); $n_d$ refers to the size of $D_t$; $A_{t-1} = \{a_i|i = 1,\cdots, t-1\}$ refers to a sequence of $t-1$ past actions sequentially assigned by the decision-maker to the client before the current time $t$, in which $a_i$ refers to the action assigned at the past time step $i$ ($i \le t-1$); and $O_{t} = \{O_{t,i} | i = 1, \cdots, t\}$ refers to a sequence of client responses to the correspondingly assigned actions during the interaction time period, in which $O_{t,i} = \{o^j | j=1,\cdots,n_r\}$ refers to a set of responses consisting of $o^j$ made by the client at time step $i$ to action $a_{i-1}$ ($i \le t$) and $n_r$ is the size of the response set. $C_t$ thus jointly captures the client's circumstances, behaviors, prior decision-making actions taken on the client, and the client responses to the actions. Accordingly, $C_t$ forms a comprehensive representation of client states, which will be further used to model the interactions with the decision-maker and quantify the effect of next-best action candidates. Further, after taking  action $a_i$, a reward value $r_{<{C_i,a_i}>}$ measures the effectiveness of $a_i$ on the client's next responses $O_{t, i+1}$. 
The larger $r_{<C_i,a_i>}$ indicates higher effectiveness. At the current time $t$, a subset of $k$ actions $\hat{A}^*_t = \{a^{j}_t | j = 1, \cdots, k\}$ are selected as the next-best actions on the client from a candidate action set $A^{*}_t$ satisfying policy constraints to achieve the top-$k$ highest rewards $\{r_{<C_t, a^{j}_t>}| j = 1, \cdots, k\}$. In practice, $k=1$ indicates that only the action associated with the highest reward is recommended, corresponding to the \textit{next-best action}. 

By empowering reinforcement learning \cite{mnih2015human,silver2016mastering,silver2017mastering} for sequential and interactive decision-making, the \textit{next-best action} corresponds to the decision action that can lead to the highest reward per  client state and to achieve the decision goal, which is learned by an action-value function $r_{\bm{\theta}}(\cdot, \cdot)$. We learn the action-value function $r_{\bm{\theta}}(\cdot, \cdot):\mathcal{C}\times\mathcal{A} \rightarrow \hat{\mathcal{R}}$, which formulates the response's reward $r_{<C_i,a_i>}$ of action $a_i$ ($a_i \in \mathcal{A}$) on the client's representation $C_i$ ($C_i \in \mathcal{C}$) at time step $i$, where $\mathcal{C}$, $\mathcal{A}$, and $\hat{\mathcal{R}}$ are the spaces of client descriptions, decision-making actions, and estimated rewards. Assuming $\mathcal{R}$ represents the space of a real reward, the personalized next-best actions $\{a^j_t|j=1,\cdots,k\}$ at time $t$ for client $c$ satisfies the following objective function:
\begin{equation}
    \begin{aligned}\label{eqn:objfun}
    & \underset{\{a^j_t|j=1,\cdots,k\}}{\text{minimize}} 
    & & Div(\hat{\mathcal{R}}||\mathcal{R}) - \sum\limits_{j=1}^{k}r_{\bm{\theta}}(C_t, a^j_t)\\
    & \text{subject to}
    & & a^j_t \in A^{*},
    \end{aligned}
\end{equation}
where $Div(\cdot||\cdot)$ is the divergence between the estimated reward space $\hat{\mathcal{R}}$ and the actual reward space $\mathcal{R}$, and $\bm{\theta}$ refers to the parameters in the action-value function $r_{\bm{\theta}}(\cdot,\cdot)$. 

The above action-value function differs from the typical reinforcement learning settings and Markovian decision processes where the action-value should be modeled as $r_{\bm{\theta}}(\cdot, \cdot):\mathcal{O} \times \mathcal{A} \rightarrow \mathcal{R}$, i.e., on the dependence between decision actions $a_t$ and client responses $O_{t,t}$ ($O_{t,t} \in \mathcal{O}$; $\mathcal{O}$ is the space of the client's responses), which only selects the action based on the current state but ignores the client's sequential behaviors in history. On the contrary, our action-value function captures the client circumstance $D_t$ and his sequences of response behaviors $O_t$ on decision-making actions $A_{t-1}$ using a comprehensive client description $C_t$ rather than client responses $O_{t,t}$ at each time step $t$. Our action-value function thus models the complex dependencies between states, between actions, and between states and actions in the sequential state-action-response-coupled sequences (Fig \ref{fig:nbaexample}), which sufficiently represent past-to-present interactions between a client and his decision-maker during sequential and interactive decision-making processes.

We further adopt empirical error minimization to learn the action-value function $r_{\bm{\theta}}(\cdot, \cdot)$ in Eq (\ref{eqn:objfun}). For a group of $n_c$ clients at time step $t$, we collect information about historical sequences of decision actions, responses, and rewards of each client $c^{(j)}$, and define the objective function below to learn the action-value function capturing the long-term dependent interactions within the client group:
\begin{equation}\label{eq:reward_obj}
    \underset{\bm{\theta}}{\text{minimize}} \sum\limits_{j = 1}^{n_c}\sum\limits_{i=1}^{t^{(j)}}l(r_{\bm{\theta}}(C_{i}^{(j)}, a_{i}^{(j)}), r_{<C_{i}^{(j)}, a_{i}^{(j)}>}),
\end{equation}
where $l(\cdot, \cdot):\mathcal{R}\times \mathcal{R} \rightarrow \mathcal{R}$ refers to a loss function that measures the difference between the real and estimated rewards, $C_{i}^{(j)}$ refers to the description of the $j$-th client at time step $i$, $a_{i}^{(j)}$ refers to the historical decision action on the $j$-th client at time step $i$, and $t^{(j)}$ refers to the maximal length of historical sequence of the $j$-th client. Our model also captures the client's behaviors within function $r_{\bm{\theta}}(\cdot,\cdot)$, which caters for personalized recommendation for each client $c^{(j)}$. Consequently, rather than only assuming the Markovian property between states, we model the long-term dependencies between client states, between decision actions, and between states and actions by jointly involving client circumstances, response behaviors to actions, and action constraints and rewards. In doing so, we capture the rich, personalized and evolving couplings and interactions in sequential, dynamic and interactive decision-making processes between individual clients in their group. After learning the action-value function $r_{\bm{\theta}}(\cdot,\cdot)$, we further learn the personalized next-best actions $\hat{A}^*_t = \{a_t^j|j=1,\cdots,k\}$ from the candidate action set $A^*_t$ by optimizing the following objective function:
\begin{equation}\label{eq:obj}
    \begin{aligned}
    & \underset{\{a^j_t|j=1,\cdots,k\}}{\text{maximize}} 
    & &  \sum\limits_{j=1}^{k}r_{\bm{\theta}}(C_t, a^j_t)\\
    & \text{subject to}
    & & a^j_t \in A^{*}_t.
    \end{aligned}
\end{equation}

For example, for the aforementioned debt collection, we model each debtor's state at time $t$ by involving the debtor's demographics, debt amount and duration, historical debt collection actions applied by the government, and response behaviors, etc. to represent the debtor's current description $C_t$, and further collect optional and sequential debt collection actions $A^{*}_t$ considerable by the government.
We aim to optimize the objective function in Eq (\ref{eq:obj}) to obtain the next-best intervention actions $\hat{A}^*_t = \{a_t^j|j=1,\cdots,k\} \subseteq A^*_t$ on each debtor.

\subsection*{Modeling the process of personalized next-best action-oriented decision-making}\label{sec:framework}

\begin{figure}[!h]
	\includegraphics[width=0.75\textwidth]{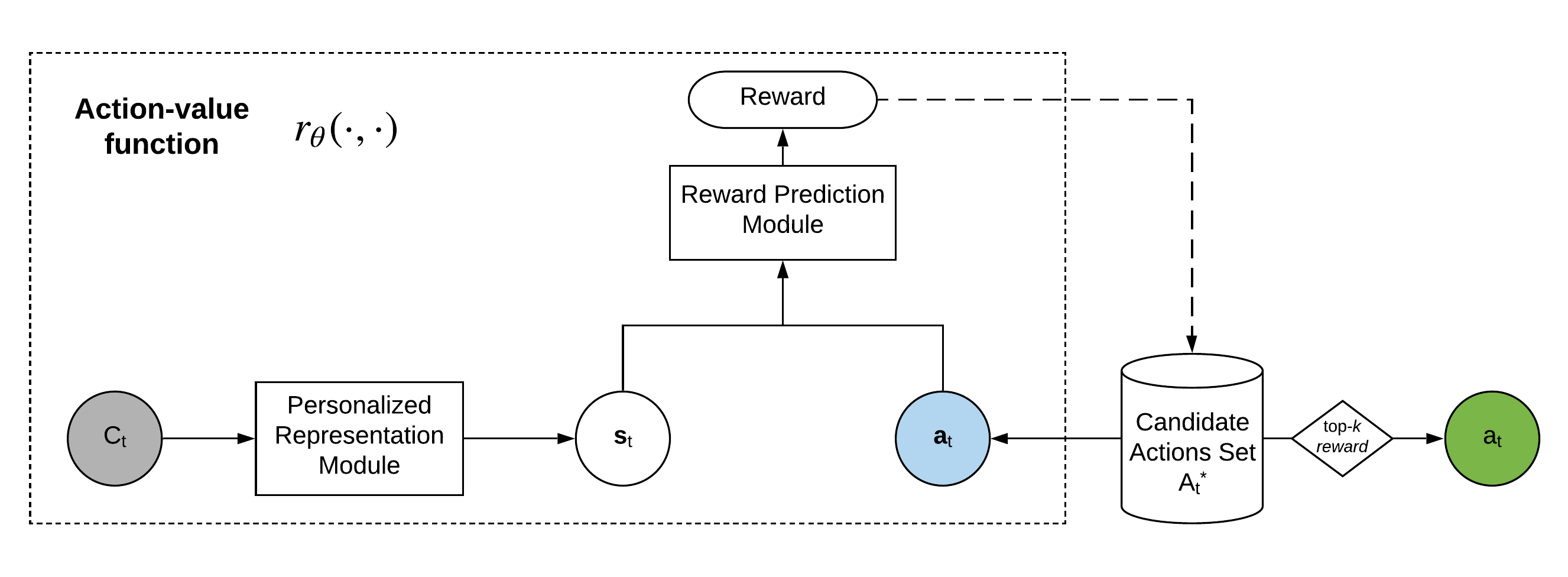}
	\caption{{\bf The framework for modeling the next-best action-oriented personalized decision-making.} $C_t$ refers to the representation describing client $c$ at time $t$, $\mathbf{s}_t$ is the vector of the client's state representation, $a_t$ refers to an action selected from the candidate action set $A^*_t$, $\mathbf{a}_t$ refers to the vector representation of action $a_t$, and $\hat{A}^*_t$ is the set of recommended next-best actions. The recommender first embeds a client's demographics, behaviors and current state to a state vector $\mathbf{s}_t$ by the personalized representation module (Fig \ref{fig:profiling}), then feeds $\mathbf{s}_t$ and $\mathbf{a}_t$ into the reward prediction module to evaluate the effectiveness of the action. The actions in the candidate set with the top-$k$ highest rewards are then recommended as the next-best actions.}
	\label{fig:framework}
\end{figure}

We model the personalized next-best action-oriented decision-making process as a personalized next-best action recommender, as shown in Fig \ref{fig:framework}. The next-best action recommender achieves the objective defined in Eq \eqref{eq:obj} in terms of two main learning tasks: (1) learning the action-value function, and (2) selecting the next-best actions. The first task learns the action-value function $r_{\bm{\theta}}(\cdot,\cdot)$, which is then used in the second task to evaluate the actions in the candidate set based on a client's behaviors and current state. Those actions with the top-$k$ highest rewards are then recommended as the next-best actions. 

Learning the action-value function is achieved by learning the personalized client representation and the action reward prediction. The personalized client representation module represents each client $C_t$ in terms of the client's demographics, behaviors and current state as a vector $\mathbf{s}_t$, which represents the client state at time $t$. The action reward prediction module further feeds $\mathbf{s}_t$ to a selected action $a_t^*$ and evaluates the action reward (i.e., effectiveness) in terms of the learned action-value function $r_{\bm{\theta}}(\cdot, \cdot)$. Those actions with the top-$k$ highest rewards are selected from the candidate action set and recommended as the next-best actions $\hat{A}_t^*$.
This design enables the candidate action set to be dynamically updated, which fits dynamic and constrained decision-making environments, where decision actions are constrained by related policies and/or environmental settings. This approach is also more efficient that other approaches such as multi-class classification-based action recommendation, because it does not need to estimate the probabilities of all possible actions (such estimation is often inefficient and may generate meaningless results in practice).

\subsection*{Personalized client representation by coupled recurrent networks}
\label{sec:personalCRN}

We represent each client description $C_t$ by a personalized client representation module. It captures the relatively stable client circumstances and the sequence of prior response behaviors to a sequence of corresponding past actions applied for decision-making up to time $t$. As a result, each client $c$ is comprehensively yet compactly represented by a state vector $\mathbf{s}_t$ at time $t$. This transforms a client's cumulative behaviors, current state and sensitivity to decision actions into a universal vector space. This personalized client representation of each client's past and current situation forms a universal yet tailored foundation to further determine different decision-making tasks on the client level and makes it benchmarkable for different clients with the same state representation. We thus can make personalized next-best action recommendation in this client representation space for each client.

Since a client's past behavior sequence reflects his personal responses and preferences to the actions taken by decision-makers in the past, different actions will be selected as the next-best ones to be taken on clients who  share similar states to fit their respective preferences and achieve the best possible reward for each client. Our method reveals the cumulative action effectiveness and the sensitivity of a client to actions by learning the complex interactions between a client's responses and assigned actions. In addition, involving a client's personal information at each time point further explains the fitness between decision actions and client circumstances. For example, debtors with different demographics and family situations likely respond differently to the same debt collection action in a government debt recovery campaign. Our approach of integrating a client's historical behavior sequence and their current personal information captures comprehensive factors affecting decision-making and is much more powerful than Markovian process models and other relevant methods. 

\begin{figure}[!h]
	\includegraphics[width=1.0\textwidth]{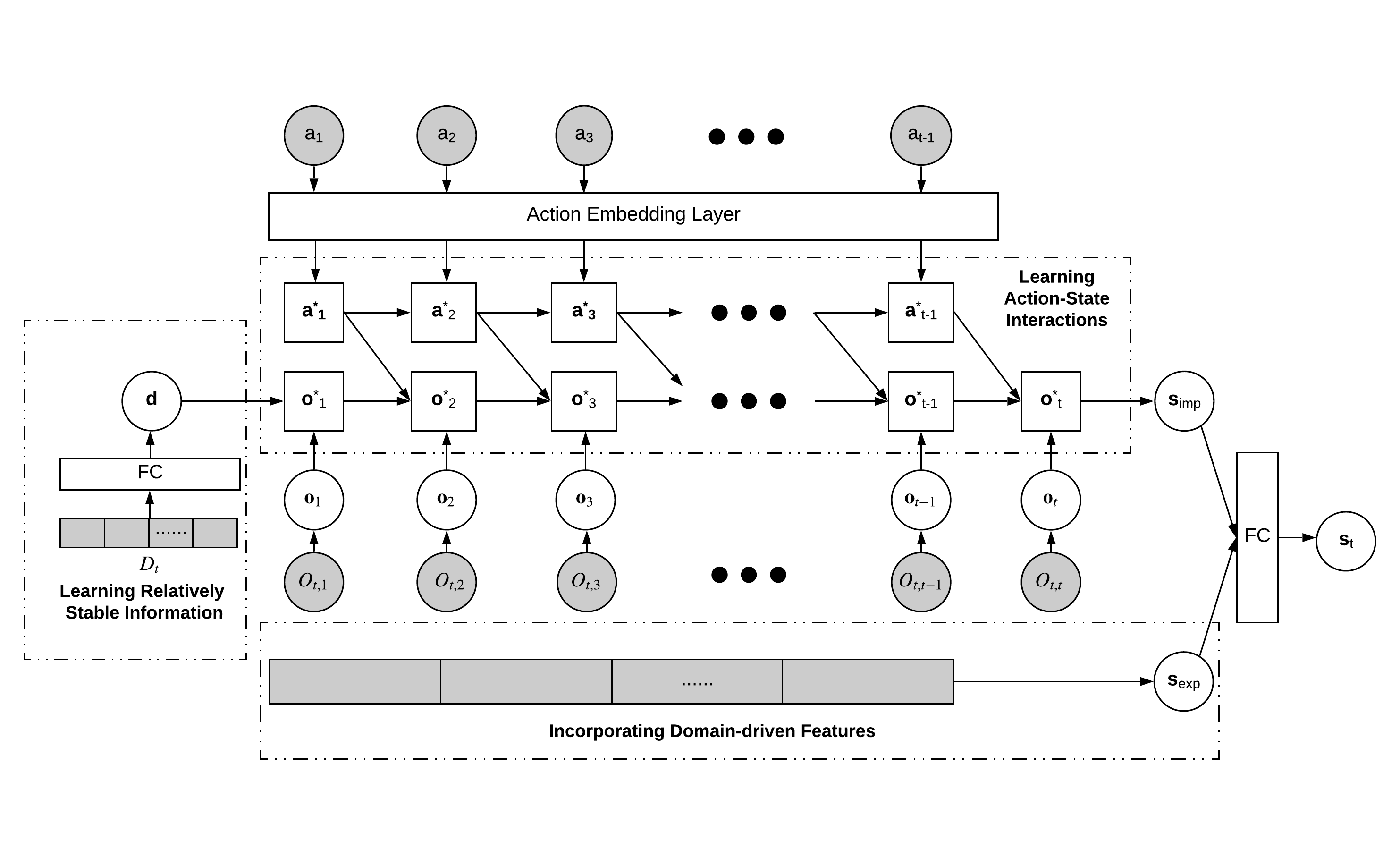}
	\caption{{\bf A reinforced coupled recurrent network to learn personalized client representation.} Given a client $c$ at current time $t$ with the description $C_t$, $O_{t,i}$ refers to the client response at past time $i$, $a_i$ is the decision action assigned to the client, $\mathbf{o}_i$ represents the vector representation of $O_{t,i}$, $\mathbf{a}_i$ represents the vector representation of $a_i$, $\mathbf{a}_i^*$ represents the hidden state corresponding to the action, $\mathbf{o}_i^*$ represents the hidden state corresponding to the client response, $\mathbf{d}$ is the transformed vector corresponding to the client's relatively stable personal information $D_t$, $\mathbf{s}_{imp}$ indicates the learned data-driven implicit features, $\mathbf{s}_{exp}$ refers to the transformed domain-driven explicit features, $\mathbf{s}_t$ is the resultant state vector representation for the client $c$, and FC refers to fully connected networks.}
	\label{fig:profiling}
\end{figure}

We learn the personalized client representation using a \textit{coupled recurrent network} (CRN, Fig \ref{fig:profiling}). Given a client tuple $C_t = <D_t, A_{t-1}, O_t>$, the decision action $a_i \in A_{t-1}$ and the set of client responses $O_{t,i}\in O_t$ at each prior time step $i$ are sequentially fed into the CRN. Initially, the client response's hidden state is extracted by a fully connected network from the client's relatively stable personal information. 
An embedding layer transforms actions described by categorical values (e.g., sending a message to a debtor) to numerical vectors. CRN embeds the client behaviors and personal information as a vector $\mathbf{s}_{imp}$, which describes the hidden state of each client at time $t$ in terms of a data-driven implicit feature since $\mathbf{s}_{imp}$ is purely generated based on the client's observable data and its characteristics by the deep network. We also extract domain-driven explicit features designed by domain experts to describe the explicit situations in the CRN and transform it to a vector $\mathbf{s}_{exp}$. Lastly, a client's current state is represented by a vector $\mathbf{s}_t$ which fuses the client's hidden state $\mathbf{s}_{imp}$ and explicit state $\mathbf{s}_{exp}$ through fully connected layers. 

\begin{figure}[!h]
	\includegraphics[width=0.65\textwidth]{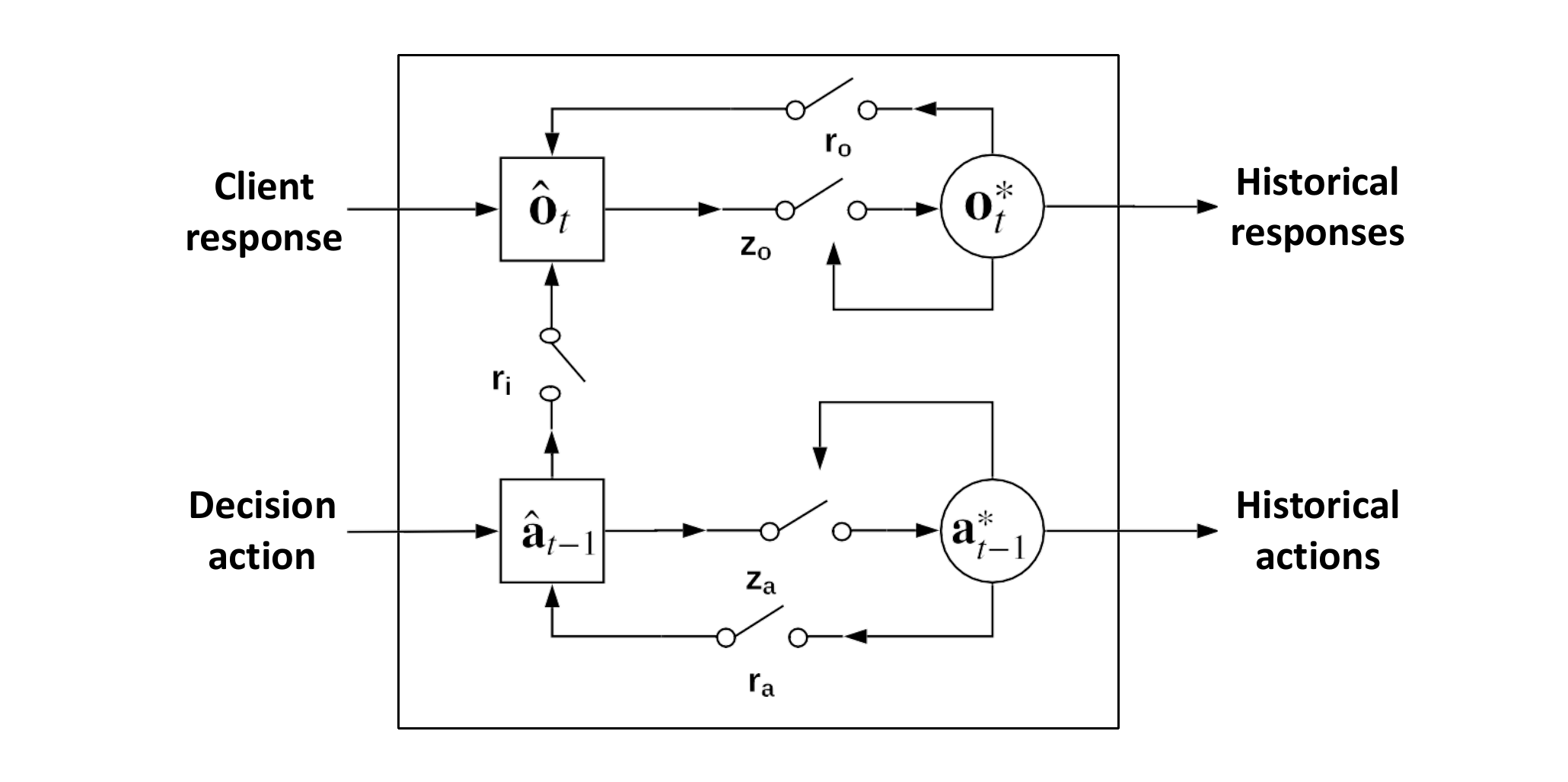}
	\caption{{\bf A coupled recurrent unit (CRU) for modeling state-action-response-coupled long-term dependencies.} $\mathbf{a}_{t-1}^*$ and $\mathbf{o}_t^*$ refer to the representation vectors of the historical sequences of actions and client responses, respectively. $\mathbf{r}_o$\ and $\mathbf{r}_a$ are two gates to control the impact of historical responses and actions on their current states. Gates $\mathbf{z}_o$ and $\mathbf{z}_a$ control the impact of current response and action states on updating the memory of their historical information respectively. $\mathbf{r}_i$ is an interaction gate to capture the dependence between a decision action and a client response.}
	\label{fig:cgru}
\end{figure}

CRN captures the complex couplings and interactions within and between the sequences of client states and decision actions in history and models client historical behaviors and interactions with the decision-maker using a coupled recurrent unit (CRU, Fig \ref{fig:cgru}). Similar to the gated recurrent unit (GRU) \cite{cho2014learning}, CRU stores the historical information in its outputs. However, there are two outputs in CRU rather than one as in GRU, which correspond to actions and responses, respectively. Specifically, the historical sequences of actions and client's responses are stored in $\mathbf{a}^*_{t-1}$ and $\mathbf{o}^*_t$, respectively. CRU adopts two gates $\mathbf{r}_o$\ and $\mathbf{r}_a$ to control the impact of historical response and action information on their current states respectively. Meanwhile, gates $\mathbf{z}_o$ and $\mathbf{z}_a$ control the impact of current states on updating the memory of historical information. In addition, CRU has an interaction gate $\mathbf{r}_i$ to capture the dependence between a decision action and a client response. With  vector representation $\mathbf{o}_t$ of the client's response $O_{t,t}$ at  time $t$ and  vector representation $\mathbf{a}_{t-1}$ of  decision action $a_{t-1}$ at  time $t-1$, the variables in CRU are calculated as follows:
\begin{equation}
    \mathbf{z}_a = \sigma(\mathbf{W}_{z_a}\mathbf{a}_{t-1} + \mathbf{U}_{z_a}\mathbf{a}_{t-2}^*),
\end{equation}
\begin{equation}
    \mathbf{r}_a = \sigma(\mathbf{W}_{r_a}\mathbf{a}_{t-1} + \mathbf{U}_{r_a}\mathbf{a}_{t-2}^*),
\end{equation}
\begin{equation}
    \mathbf{z}_o = \sigma(\mathbf{W}_{z_o}\mathbf{o}_t + \mathbf{U}_{z_o}\mathbf{o}_{t-1}^*),
\end{equation}
\begin{equation}
    \mathbf{r}_o = \sigma(\mathbf{W}_{r_o}\mathbf{o}_t + \mathbf{U}_{r_o}\mathbf{o}_{t-1}^*),
\end{equation}
\begin{equation}
	\mathbf{r}_i = \sigma(\mathbf{W}_{i}\mathbf{a}_{t-1} + \mathbf{U}_{i}\mathbf{o}_{t-1}^*),
\end{equation}
\begin{equation}
	\hat{\mathbf{a}}_{t-1} = tanh(\mathbf{W}_{a}\mathbf{a}_{t-1} + \mathbf{U}_{a}(\mathbf{r}_a\circ \mathbf{a}_{t-2}^*)),
\end{equation}
\begin{equation}
    \hat{\mathbf{o}}_t = tanh(\mathbf{W}_{o}\mathbf{o}_{t} + \mathbf{U}_{o}(\mathbf{r}_o\circ \mathbf{o}_{t-1}^*) + \mathbf{I}_{o}(\mathbf{r}_i\circ \hat{\mathbf{a}}_{t-1})),
\end{equation}
\begin{equation}
	\mathbf{a}_{t-1}^* = (\mathbf{1}_a-\mathbf{z}_a)\circ \mathbf{a}_{t-2}^* + \mathbf{z}_a\circ \hat{\mathbf{a}}_{t-1},
\end{equation}
\begin{equation}
    \mathbf{o}_t^* = (\mathbf{1}_o-\mathbf{z}_o)\circ \mathbf{o}_{t-1}^* + \mathbf{z}_o\circ \hat{\mathbf{o}}_t,
\end{equation}
where $\sigma(\cdot)$ is the sigmoid function, $tanh(\cdot)$ is the hyperbolic tangent function, $\circ$ refers to the Hadamard product, $\mathbf{1}_a$ and $\mathbf{1}_o$ are vectors with all elements as 1 and with a $n_a \times 1$ dimension and a $n_o \times 1$ dimension, respectively, $\mathbf{W}_{z_a}, \mathbf{W}_{r_a}, \mathbf{W}_{i}, \mathbf{W}_{a}, \mathbf{U}_{z_a}, \mathbf{U}_{r_a}$, and $\mathbf{U}_{a}$ are learnable matrices with a $n_a \times n_a$ dimension, $\mathbf{W}_{z_o}, \mathbf{W}_{r_o}, \mathbf{W}_{o}, \mathbf{U}_{z_o}, \mathbf{U}_{r_o}$ and $\mathbf{U}_{o}$ are learnable matrices with a $n_o \times n_o$ dimension,  $\mathbf{U}_{i}$ is a learnable matrix with a $n_a \times n_o$ dimension, and $\mathbf{I}_{o}$ is a learnable matrix with a $n_o \times n_a$ dimension, $n_o$ is the dimension of response vector representation $\mathbf{o}$, and $n_a$ is the dimension of action embedding $\mathbf{a}$.

\begin{figure}[!h]
	\includegraphics[width=0.75\textwidth]{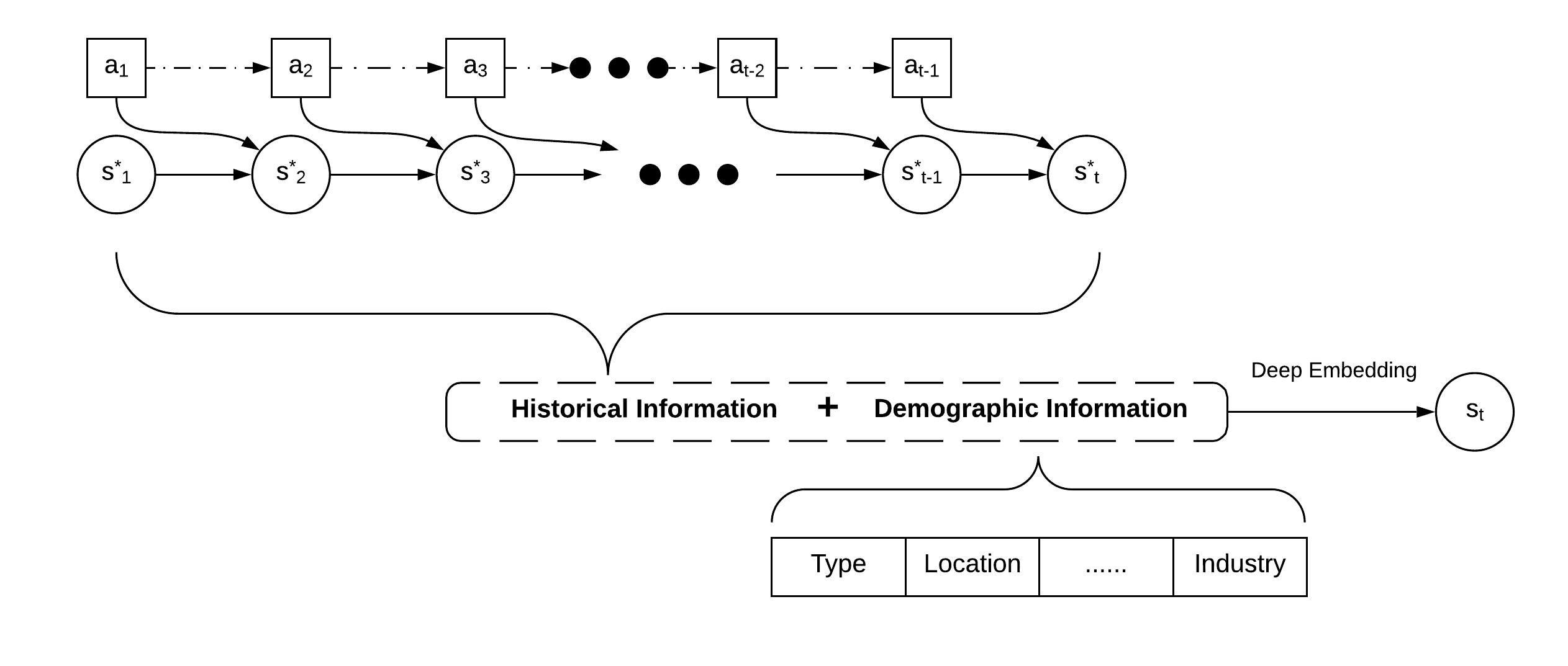}
	\caption{{\bf An example of representing three clients by the reinforced coupled recurrent network.} Three debtors with different demographics and past response behaviors to the same decision actions are represented in three vectors.}
	\label{fig:example}
\end{figure}

As a result, each client is comprehensively represented in terms of his circumstances, past decision actions received, past responses to the actions, and domain-driven factors considered in the decision-making process. For all clients, a personalized representation (see an example in Fig \ref{fig:example}) is learned for each of them. The learned representations differ from or are similar to each other, corresponding to the similarity between their demographics and responses to actions. This provides a universal, comprehensive and benchmarkable representation to further conduct personalized decision-making.

\subsection*{Reward prediction of next-best actions on client states}
\begin{figure}[!h]
	\includegraphics[width=0.45\textwidth]{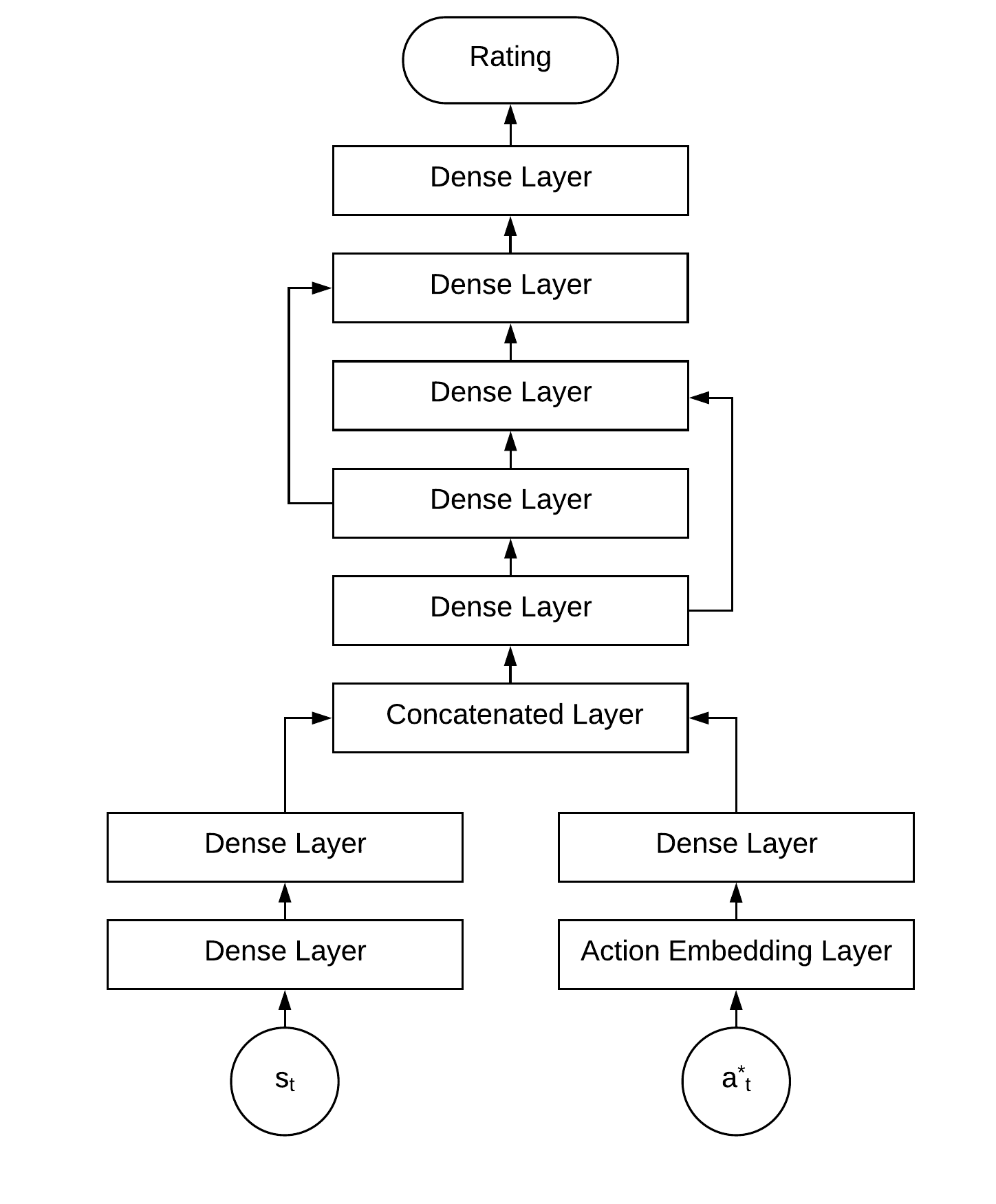}
	\caption{{\bf Reward prediction for the next-best action on a client's state.} The reward (rating) of an action is predicted by residual networks corresponding to a client's state.}
	\label{fig:rpm}
\end{figure}

We further measure the reward of each decision action on a client state using a reward prediction module (Fig \ref{fig:rpm}), which is built on a residual network. 
The above learned client state representation vector $\mathbf{s}_t$ and an action $a_t^j$ selected from a set of candidate actions $A^*_t$ that satisfy decision-making policy constraints are input into the reward prediction module. The candidate action $a_t^j$ is first embedded through an action embedding layer (the same as the action embedding layer in the personalized client representation module) to $\mathbf{a}_t^{j}$. Further, this embedded action is concatenated with the client state representation vector $\mathbf{s}_t$ as the input of the following three-layer residual network. The last layer of the residual network predicts the reward $r_{\bm{\theta}}(C_t, a_t^{j})$ of each input action $a_t^j$ corresponding to the target client state $C_t$.

The residual network-based reward prediction module shows unique strengths in efficiently modeling large-scale sequential decision-making actions. First,  reward prediction is efficient in processing a large number of states and actions since it learns a common reward prediction model for different clients. Given a client state representation, it  efficiently predicts the reward values for different actions. Second, modeling complexity can be automatically controlled since the residual network structure is embedded with a potential bypass from low-level information to high-level information. When the input data involves hierarchical patterns, the high-level features will be learned for the final prediction. For data with simple patterns, the low-level features will make a direct contribution to the final prediction. This reduces over-fitting in reward prediction and enables personalized client representation to be well learned to capture heterogeneous client behaviors, which are embedded in a common space for further decision-making tasks.

The next-best action recommendation module assesses the learned reward  $r_{\mathbf{\theta}}(\cdot,\cdot)$ associated with each action in the candidate set $A^*_t$ for client $c$ at  time step $t$ to judge the effectiveness of taking action for decision-making. Those actions with the top-$k$ ($k$ is a hyperparameter to be determined by decision makers) highest rewards are recommended as the next-best actions $\hat{A}^*_t \subseteq A^*_t$ for the client. 

\subsection*{Strategies to learn from hierarchical imbalanced action-response interactions}\label{sec:imbalancedLearning}

Real-life data often presents imbalanced distributions \cite{krawczyk2016learning}. In our case study of five-year debt collection data, we find it highly imbalanced and hierarchical across the attributes, attribute values, domain-driven rewards, and reward levels (Table \ref{tab:utility_distribution}). With respect to the actions, their frequency distribution is extremely imbalanced, which we call \textit{action imbalance}. Some commonly taken actions may appear thousands of times more than other rarely taken actions. Regarding the client interactions, the counts of interactions between actions and clients are imbalanced, resulting in \textit{client interaction imbalance}. For example, a small fraction of the client cohort may involve a large proportion of interactions. With regard to the reward of actions, most of the reward values given by domain experts to actions may be 0, leading to \textit{reward imbalance}. Lastly, the action effectiveness is different, where a small number of actions are very strong and effective, thus they always generate a high reward, resulting in \textit{action effectiveness imbalance}.

These hierarchical imbalanced distributions in actions, interactions, rewards and action effects bring a significant challenge to the personalized recommendation of next-best actions. Action imbalance makes the model sensitive to those actions with high frequency but insensitive to the rarely appearing actions. This is caused by the model parameters that are trained predominantly by samples with high-frequency actions in the training phase if the imbalance is not catered for. The client interaction imbalance also affects the training of CRN. Since the sequence lengths of past client behaviors and decision actions are both short in most cases, it is difficult for CRN to effectively capture the long-term dependencies in those few but long historical sequences. Further, reward imbalance induces the reward prediction of the model to be 
$0$. This results in most prediction results being $0$, hence the model cannot generate the next-best actions. In addition, the action effectiveness imbalance also results in the model consistently selecting those highly effective actions (which are usually tough actions) by prediction, which tends to recommend tough actions at all times for all clients. However, such recommendations mostly violate government service policies and constraints. In addition, the various imbalances are mixed with each other in the action-response interaction sequences, further increasing modeling difficulty. Consequently, the imbalanced distributions at different aspects bring significant but different challenges to personalized next-best action modeling. 

Accordingly, we propose several strategies to improve  CRN training (Section \ref{sec:personalCRN}) and tackle the challenges brought by the hierarchical imbalances in action-response interactions. The key idea behind these strategies is to introduce explicit knowledge to regulate the implicit learning of multiple sequences and their dependencies in CRN (see the section on personalized client representation). Specifically, the various imbalances are first statistically quantified; then, the statistic information is used to sample the training data, weight the importance of samples, and adjust the effect on reward prediction loss. The respective strategies to tackle the imbalance at different aspects  are as follows.
\begin{itemize}
	\item \textit{Action imbalance}: Setting the weight of client $c$ with  action $a^i$ as 
	\begin{equation}
	w_c^i = \frac{\exp({1}/{f_{a^i}})}{\sum_{j = 1}^{m}\exp({1}/{f_{a^j}})},
	\end{equation}
	where $f_{a^j}$ is the frequency of  action $a^j$, and $m$ is the total number of actions. To reflect  action imbalance in the loss function (Eq \eqref{eq:obj}), the loss value on the client is multiplied by $w_c^i$ for  backward gradient propagation.
	\item \textit{Client interaction imbalance}: Sampling the training data with probabilities $\{p^i|i = 1,\cdots,n_c\}$ for all $n_c$ clients in each batch, $p^i$ is the sampling probability of the $i$-th client and is calculated as
	\begin{equation}
	p^i = \frac{\exp(l^i)}{\sum_{j=1}^{n_c}\exp(l^j)},
	\end{equation}
	where $l^j$ is the length of historical information of the $j$-th client and $n_c$ is the number of clients.
	\item \textit{Reward imbalance}: Setting the weight of the reward $r_{<C_t, a_t>}$ to action $a_t$ on client $C_t$ as 
	\begin{equation}
	w_r(r_{<C_t, a_t>}) = \tanh(r_{<C_t, a_t>} + 0.1).
	\end{equation}
	The loss value (Eq \eqref{eq:obj}) of the client with reward $r_{<C_t, a_t>}$ is multiplied with $w_r$, and only the top-$k$ largest loss values in a batch are selected for backward gradient propagation.
	\item \textit{Action effectiveness imbalance}: Adjusting reward $r_{<C_t, a_t>}$ in training samples as 
	\begin{equation}
	r_{<C_t, a_t>}^* = r_{<C_t, a_t>}/t^2,
	\end{equation}
	where $t$ is the time duration (i.e., the current time step) when action $a_t$ is assigned.
\end{itemize}

\section*{Results}

\subsection*{The pilot settings and characteristics}\label{sec:casestudy}
A backtesting of our personalized next-best action recommendation was conducted on five-year (2012-2017) debt collection data in a major Australian government agency. A subset of 5-year debt-related data from the government was used, which comprises 61,361 clients, 10 selected debt collection actions, and 66,126 client response-government action sequences in a total of 111,514 debt transactions. The data comprises attributes about client demographics and circumstances, the debt amount and duration at each time point associated with a debtor, a list of optional debt collection actions and their application policy constraints, a sequence of historical actions taken by the government on a debtor to recover the debt at each time point, the corresponding client response behavior to each debt collection action, and the time information associated with debt cases, responses and actions. 

In debt collection, those actions that likely bring about faster and more debt recovery are deemed as  high reward. Debt collection experts rate the reward associated with each action on the debtor population (rather than individual debtors). Accordingly, we categorize all optional actions into two categories: (1) the low-reward action group  where actions receive reward less than 0.5, and (2) the high-reward action group where actions receive reward larger than 0.5. The corresponding reward distribution of 10 selected debt collection actions (annotated for privacy consideration) is shown in Table \ref{tab:utility_distribution}, where the distribution of actions and their rewards over the five years is highly imbalanced. The most frequent action is Action 6 (A6) which appeared 62,263 times, while the least frequent action is Action 2 (A2) which only appeared 390 times. The length distribution of historical action sequences on each debtor is also imbalanced. Only 50\% of clients had their action sequence length larger than 4. In addition, the domain-driven rewards given to these debt collection actions are also imbalanced, the mean reward of all actions is under 0.32, and the highest reward given to all actions equals 1. These show the need for handling the hierarchical imbalances with our strategies proposed in Section \ref{sec:imbalancedLearning}.

We randomly split the data into training, validation and testing sets in proportions of 70\%, 10\%, 20\% respectively. Due to resource constraints in the pilot, the government only selected a proportion of debtors from the entire debtor pool to apply the intervention actions recommended by our method. We calculated the average domain-driven reward given by the debt collectors of the 10\% highest predicted reward by our CRN model and reported it as our modeling performance.
This result was agreed by the debt collectors to indicate how much percentage of debts can be deducted on average if the debt intervention was based on the next-best actions recommended by our model. For privacy reasons, we cannot report the government information or any details about the debtors and debt collectors in the pilot and cannot directly report the average debt deduction percentage incurred by our recommendations in comparison to that driven by the government's rule-based action selection strategies. Instead, we report the reward lift and error reduction made by our model recommendations in comparison with the domain-driven debt collection rules.

\begin{table}[ht] 
	\caption{The distribution of reward of 10 actions specified by debt collection experts. 10 actions were chosen by the government. Count(Reward $< 0.5$) measures the number of action occurrences in the five years which have been given reward lower than 0.5 by debt collectors. Count(Reward $>=0.5$) refers to the number of action occurrences with reward larger than 0.5. Reward Mean refers to the mean of all rewards per action. Reward Std refers to the standard deviation of the rewards per action. High-reward Proportion refers to the percentage of the rewards per action.}
	\small
	\begin{center}
		\begin{tabular}{|l|l|l|l|l|l|l|}
			\toprule
			Action ID & Count & Count(Reward $< 0.5$) & Count(Reward $>=0.5$) & Reward Mean & Reward Std & High-reward Proportion \\ 
			\midrule
			A1 & 1,225 & 1,196 & 29 & 0.065 & 0.145 & 2.42\% \\ 
            A2 & 390 & 349 & 41 & 0.132 & 0.259 & 11.75\% \\ 
            A3 & 13,592 & 12,640 & 952 & 0.125 & 0.221 & 7.53\% \\ 
            A4 & 1,020 & 800 & 220 & 0.229 & 0.34 & 27.50\% \\ 
            A5 & 1,384 & 1,340 & 44 & 0.057 & 0.158 & 3.28\% \\ 
            A6 & 62,263 & 49,473 & 12,790 & 0.223 & 0.333 & 25.85\% \\ 
            A7 & 15,403 & 13,090 & 2,313 & 0.186 & 0.294 & 17.67\% \\
            A8 & 3,289 & 3,084 & 205 & 0.097 & 0.205 & 6.65\% \\ 
            A9 & 904 & 643 & 261 & 0.311 & 0.355 & 40.59\% \\ 
            A10 & 12,044 & 10,753 & 1,291 & 0.159 & 0.262 & 12.01\% \\
            Total & & 93,368 & 18,146 & & & 19.43\% \\ 
			\midrule
			\bottomrule
		\end{tabular}
	\end{center}
	\label{tab:utility_distribution}
\end{table}

\subsection*{Baseline methods}\label{subsec:baselines}

We test our CRN model against (1) domain-driven rules i.e. the debt collection rules defined by the debt collection experts, (2) variants of three state-of-the-art deep models with modifications to cater for the next-best action recommendation: Google's wide-and-deep (WD) model, LSTM and GRU-based RNNs, and (3) the combination of wide-and-deep model with RNN strategies. Specifically, the domain-driven rules were taken by the government, where debt collection actions were taken according to the government's debt collection policies and constraints defined by debt collection experts. Such domain-driven rule-based action-taking method reflects the best practice in the debt collection business and was taken as best practice, thus we treat it as the baseline to evaluate the effectiveness and business impact of our model recommendations. Second, the WD model was shown to achieve  state-of-the-art results in recommendation \cite{cheng2016wide}. It reflects the performance of the state-of-the-art Markov decision process, and we revise it to learn decision rules based on the current state of a client. Third, the LSTM and GRU-based RNNs are shown to be effective in learning long-term dependencies. We embed historical client states into LSTM and GRU to transform a non-Markovian decision process to a Markovian decision process. They serve as the performance benchmark of the state-of-the-art Markovian decision process learning. Lastly, we combine the WD RNN with LSTM and GRU to take advantage of the two advanced deep modeling mechanisms: residual network (Res) and multiple layers (Multi), to form the best possible non-Markovian decision process learners: WD\_LSTM, WD\_GRU, WD\_Res\_LSTM, WD\_Multi\_LSTM, WD\_Res\_GRU, WD\_Multi\_GRU. They reflect the best possible performance we may achieve by hybridizing the state-of-the-art achievements in deep learning.

We empirically evaluate the performance of the proposed personalized next-best action recommender CRN in terms of the following aspects:
(1) Ability to reveal whether our model can effectively predict an accurate reward value; (2) Business impact to demonstrate whether the recommended next-best actions can lead to an estimated high reward for business in practice; and (3) Scalability to reflect whether CRN is scalable for handling a large amount of data.

In our experiments, CRN represents each client's demographic features (e.g., client type, address, and industry sector, etc.) to form the initial states of CRU. This solves the cold-start problem in decision-making by assuming that  clients with similar demographic features likely share similar behaviors. Our model uses the \textit{ReLU} activation function \cite{hahnloser2000digital} for nonlinear mapping and has a batch-normalization layer after all non-linear layers. All multi-layer perceptron (MLP) networks in our model have three layers. We train the CRN model using the \textit{Adam} algorithm \cite{kingma2014adam} with a batch size of 128.

\subsection*{Recommendation of next-best actions for each client }
\label{sec:nbarecommder}

We applied the recommended next-best actions for five-year debt collection. As shown in Table \ref{tab:topklift}, our reward prediction module  achieves 2.1942 total average reward lift ($total\_avg$) and 2.4954 action average reward lift ($action\_avg$) in comparison with 2.1089 ($total\_avg$) and 2.2049 ($action\_avg$) by Google's best-performing WD model, leading to a 4.04\% and 13.18\% improvement, respectively in recommending 10 next-best actions that satisfy the policy constraints for debt collection. By applying a hierarchical imbalanced training strategy (discussed in the Method section) on the CRN for reward prediction, our method  achieves a reward lift of 2.5569 ($total\_avg$) and 3.4599 ($action\_avg$), which is 21.24\% and 56.92\% better than the total average and action average reward lift made by the WD model.

\begin{table}[ht] 
\caption{Average reward lift of the reward made by 10 actions recommended by 11 deep models over that by the government debt collection rules. A1 to A10 are 10 actions (Table \ref{tab:utility_distribution}) selected from five-year debt collection data by the government. CRN and CRN\_IMB are our methods, WD, LSTM, WD\_LSTM, WD\_Res\_LSTM, WD\_Multi\_LSTM, GRU, WD\_GRU, WD\_Res\_GRU and WD\_Multi\_GRU are baseline deep models. $\Delta$\_IMB and $\Delta$ refer to the improvement percentage made by CRN\_IMB and CRN in comparison with the best competitors, respectively.}
\centering
\scalebox{0.75}
{\begin{tabular}{|l|l|l|l|l|l|l|l|l|l|l|l|l|}
\toprule
Model & A1  & A2  & A3  & A4  & A5 & A6  & A7  & A8 & A9  & A10  & Total\_Avg & Action\_Avg \\
\midrule
CRN\_IMB & 5 & 4 & 3.0534 & 2.8752 & 6.8 & 2.1415 & 2.6984 & 3.3567 & 1.6772 & 2.9969 & 2.5569 & 3.4599 \\ 
CRN & 2.1957 & 3.5383 & 2.2068 & 2.6616 & 3.216 & 2.074 & 2.326 & 2.6277 & 1.7654 & 2.3425 & 2.1942 & 2.4954 \\ 
WD & 2.604 & 1.5992 & 2.0979 & 2.2798 & 3.2239 & 1.9824 & 2.2629 & 2.6967 & 0.9899 & 2.312 & 2.1089 & 2.2049 \\ 
LSTM & 0.9722 & 1.0987 & 0.9391 & 0.974 & 1.1272 & 1.0159 & 0.897 & 1.1097 & 1.1024 & 1.0847 & 1.0013 & 1.0321 \\ 
WD\_LSTM & 2.0471 & 1.2731 & 1.9709 & 2.4755 & 2.2217 & 1.8129 & 2.0816 & 2.1909 & 1.1405 & 2.105 & 1.9198 & 1.9319 \\ 
WD\_Res\_LSTM & 1.7247 & 0.8219 & 1.7007 & 1.9816 & 2.4985 & 1.8164 & 1.9851 & 2.0921 & 0.8285 & 1.967 & 1.8488 & 1.7416 \\ 
WD\_Multi\_LSTM & 1.684 & 1.0468 & 1.6591 & 1.774 & 1.6924 & 1.7083 & 1.671 & 2.1678 & 1.2222 & 1.8098 & 1.7161 & 1.6435 \\
GRU & 0.5783 & 0.0865 & 0.9852 & 1.1201 & 1.5022 & 0.9154 & 0.861 & 0.9463 & 1.0347 & 1.0416 & 0.9345 & 0.9071 \\ 
WD\_GRU & 1.0049 & 0.6397 & 1.3454 & 1.7369 & 2.1271 & 1.6489 & 1.6049 & 2.1562 & 0.665 & 1.6602 & 1.611 & 1.4589 \\
WD\_Res\_GRU & 1.4488 & 1.1333 & 1.7364 & 1.3479 & 2.2259 & 1.6932 & 1.7091 & 1.9582 & 1.2507 & 1.8869 & 1.7248 & 1.6391 \\
WD\_Multi\_GRU & 1.6329 & 1.8399 & 1.9114 & 1.7949 & 1.8781 & 1.8206 & 2.0276 & 1.7613 & 1.0508 & 2.2347 & 1.8959 & 1.7952 \\ 
\midrule
$\Delta$\_IMB & 92.01\% & 117.40\% & 45.55\% & 16.15\% & 110.92\% & 8.03\% & 19.25\% & 24.47\% & 34.10\% & 29.62\% & 21.24\% & 56.92\% \\
$\Delta$ & -15.68\% & 92.31\% & 5.19\% & 7.52\% & -0.25\% & 4.62\% & 2.79\% & -2.56\% & 41.15\% & 1.32\% & 4.04\% & 13.18\% \\
\bottomrule
\end{tabular}}
\label{tab:topklift}
\end{table}

In the pilot, those actions with an estimated reward larger than 0.5 were applied as an intervention with their debtors for faster, less costly and more debt collection. By comparing the domain-driven reward given by the debt collectors, we evaluate the precision of CRN-recommended actions in terms of calculating the percentage of domain-driven high-reward actions that CRN also predicts as high-reward ones. The results in Table \ref{tab:high_utility_precision} show that CRN results in 2.6465 (\textit{total\_avg}) and 3.2799 (\textit{action\_avg}) lift, which is 5.38\% and 11.74\% better than the best-performing WD model. CRN\_IMB further shows that CRN improves \textit{action\_avg} to 3.3816, which is 15.20\% better than the WD model. Our method largely improves the precision for those actions rarely applied in business (e.g., A2 which only appeared 100 times in five years), which are shown to be more effective for some debtors.

\begin{table}[ht] 
\caption{Precision lift of the precision achieved by 10 actions recommended by 11 deep models over that by government debt collection rules. $\Delta$\_IMB and $\Delta$ refer to the improvement percentage of our CRN\_IMB and CRN models in comparison with the best competitors, respectively.}
\centering
\scalebox{0.75}{\begin{tabular}{|l|l|l|l|l|l|l|l|l|l|l|l|l|}
\toprule
Model & A1 & A2  & A3  & A4  & A5  & A6 & A7  & A8  & A9  & A10  & Total\_Avg & Action\_Avg \\ 
\midrule
CRN\_IMB & 5.0000 & 5.0000 & 2.9824 & 2.4213 & 5.4400 & 2.2071 & 2.7393 & 3.2107 & 1.6772 & 3.1382 & 2.5833 & 3.3816 \\ 
CRN & 4.1667 & 4.0000 & 3.6215 & 2.5726 & 5.4400 & 2.2123 & 2.7666 & 3.5026 & 1.5482 & 2.9686 & 2.6465 & 3.2799 \\ 
WD & 4.1667 & 2.0000 & 2.9469 & 2.5726 & 4.7600 & 2.1127 & 2.7121 & 3.9404 & 1.0321 & 3.1099 & 2.5114 & 2.9353 \\ 
LSTM & 0.8333 & 1.0000 & 0.7456 & 0.9080 & 1.3600 & 1.0406 & 0.7632 & 1.3135 & 1.1611 & 1.1026 & 0.9814 & 1.0228 \\
WD\_LSTM & 2.5000 & 1.0000 & 2.9469 & 2.7239 & 3.4000 & 1.9345 & 2.4531 & 3.5026 & 1.2902 & 2.7424 & 2.2886 & 2.4494 \\
WD\_Res\_LSTM & 2.5000 & 1.0000 & 2.4498 & 2.1186 & 4.0800 & 1.9135 & 2.2623 & 2.7729 & 1.0321 & 2.6576 & 2.1589 & 2.2787 \\ 
WD\_Multi\_LSTM & 1.6667 & 2.0000 & 1.2090 & 1.2143 & 1.1667 & 1.1041 & 1.1988 & 1.4211 & 1.0000 & 1.1702 & 1.1561 & 1.3145 \\ 
GRU & 0.0000 & 1.0000 & 0.9941 & 0.9080 & 2.0400 & 0.8729 & 0.7496 & 1.0216 & 1.0321 & 0.8482 & 0.8785 & 0.9466 \\ 
WD\_GRU & 0.8333 & 0.0000 & 1.6687 & 1.6646 & 3.4000 & 1.7982 & 1.7444 & 2.9188 & 0.6451 & 2.1487 & 1.8392 & 1.6822 \\ 
WD\_Res\_GRU & 0.8333 & 1.0000 & 2.5918 & 1.3200 & 2.7200 & 1.8087 & 1.8262 & 2.7729 & 1.2902 & 2.4031 & 1.9885 & 1.8608 \\
WD\_Multi\_GRU & 2.5000 & 2.0000 & 2.8049 & 0.9673 & 2.7200 & 1.9161 & 2.3577 & 2.4810 & 1.2902 & 3.0251 & 2.2346 & 2.3062 \\
\midrule
$\Delta$\_IMB & 20.00\% & 150.00\% & 1.20\%&-11.11\% & 14.29\% & 4.47\% & 1.00\%& -18.52\% & 30.00\% & 0.91\% & 2.86\%&15.20\% \\ 
$\Delta$ & 0.00\% & 100.00\% & 22.89\% & -5.55\%& 14.29\% & 4.71\% & 2.01\% & -11.11\% & 20.00\% & -4.54\% & 5.38\% & 11.74\% \\ 
\bottomrule
\end{tabular}}
\label{tab:high_utility_precision}
\end{table}

We evaluate  CRN effectiveness w.r.t. the mean squared error (MSE, Table \ref{tab:mse}) of recommendations of next-best actions for five-year debt collection, which measures the difference between the domain-driven reward given by debt collection experts and the reward predicted by CRN for each action in the 10 action candidates. CRN recommendations achieve the best overall MSE results, i.e.,  \textit{total\_avg} at 0.0777 and \textit{action\_avg} of 0.0613, and CRN makes a 3.24\% and 7.26\% improvement over the best-performing WD model in terms of \textit{total\_avg} and \textit{action\_avg}, respectively. 

\begin{table}[ht] 
\caption{The reward mean squared error (MSE) per action between the reward made by the domain-driven debt collection rules and that recommended by 10 deep models. A1 to A10 are 10 actions (Table \ref{tab:utility_distribution}) selected from five-year debt collection data by the government. CRN is our method, WD, LSTM, WD\_LSTM, WD\_Res\_LSTM, WD\_Multi\_LSTM, GRU, WD\_GRU, WD\_Res\_GRU and WD\_Multi\_GRU are baseline deep models. $\Delta$ refers to the improvement percentage of the CRN compared with the best-performing method. The lower MSE indicates the better CRN performance. \textit{action\_avg} shows the unweighted average MSE and \textit{total\_avg} is the weighted average MSE of each method. The weighted average is calculated in terms of the number of actions.}
\centering
\scalebox{0.75}{\begin{tabular}{|l|l|l|l|l|l|l|l|l|l|l|l|l|}
\toprule
Model & A1  & A2  & A3  & A4  & A5 & A6  & A7  & A8 & A9  & A10  & Total\_Avg & Action\_Avg \\
\midrule
CRN & 0.0266 & 0.055 & 0.0462 & 0.094 & 0.0222 & 0.0937 & 0.0733 & 0.0384 & 0.1077 & 0.056 & 0.0777 & 0.0613 \\ 
WD & 0.0271 & 0.0631 & 0.0491 & 0.1038 & 0.0263 & 0.0963 & 0.076 & 0.0384 & 0.1245 & 0.0565 & 0.0803 & 0.0661 \\
LSTM & 0.1219 & 0.1315 & 0.1129 & 0.1411 & 0.1286 & 0.131 & 0.1201 & 0.1216 & 0.1256 & 0.1166 & 0.1253 & 0.1251 \\ 
WD\_LSTM & 0.2361 & 0.2395 & 0.2167 & 0.2188 & 0.2539 & 0.2163 & 0.2146 & 0.2352 & 0.1757 & 0.2108 & 0.2165 & 0.2218 \\ 
WD\_Res\_LSTM & 0.2188 & 0.2333 & 0.2187 & 0.2128 & 0.2363 & 0.2091 & 0.2078 & 0.2192 & 0.1776 & 0.2099 & 0.2108 & 0.2143 \\ 
WD\_Multi\_LSTM & 0.2429 & 0.2485 & 0.2203 & 0.2215 & 0.2616 & 0.2177 & 0.2161 & 0.2417 & 0.177 & 0.212 & 0.2185 & 0.2259 \\ 
GRU & 0.1011 & 0.1139 & 0.0957 & 0.1324 & 0.1035 & 0.1215 & 0.1076 & 0.103 & 0.1243 & 0.1021 & 0.1134 & 0.1105 \\ 
WD\_GRU & 0.2299 & 0.2368 & 0.2211 & 0.2174 & 0.2417 & 0.213 & 0.2106 & 0.2261 & 0.1798 & 0.2174 & 0.2149 & 0.2194 \\
WD\_Res\_GRU & 0.2301 & 0.2384 & 0.2245 & 0.2168 & 0.2493 & 0.2142 & 0.2119 & 0.2304 & 0.1777 & 0.2156 & 0.2162 & 0.2209 \\ 
WD\_Multi\_GRU & 0.228 & 0.2354 & 0.2196 & 0.2195 & 0.2443 & 0.2157 & 0.2131 & 0.2279 & 0.1795 & 0.2136 & 0.2162 & 0.2197 \\
\midrule
$\Delta$ & 1.85\% & 12.84\% & 5.91\% & 9.44\% & 15.59\% & 2.70\% & 3.55\% & 0.00\% & 13.35\% & 0.88\% & 3.24\% & 7.26\% \\
\bottomrule
\end{tabular}}
\label{tab:mse}
\end{table}

\begin{figure}[!h]
	\includegraphics[width=0.5\textwidth]{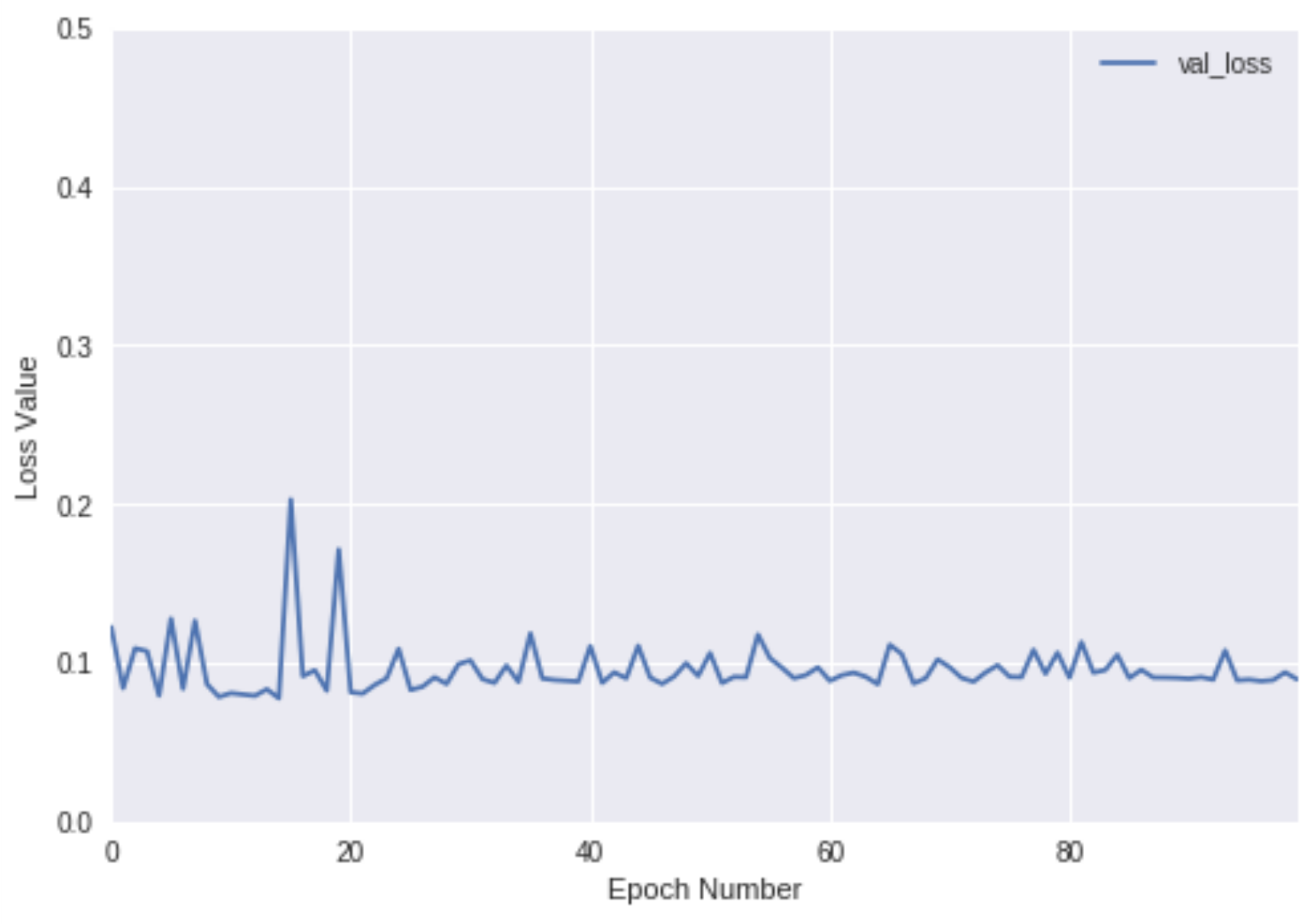}
	\caption{{\bf CRN convergence w.r.t. loss value on the validation debt collection data.} The X-axis refers to the number of epochs, and the Y-axis refers to the loss value of the CRN objective function (Eq (\ref{eq:obj})).}
	\label{fig:convergence}
\end{figure}

We further test our CRN model to show it can efficiently model large-scale client-decision-maker interactions, as shown in Fig \ref{fig:convergence}. In our test environment (Section \ref{sec:casestudy}), CRN converges within 20 epochs, and the mean computational cost in each epoch is around 2 minutes in our testing environment. These empirical results show that CRN can be applied to large-scale interaction data and problems.

\section*{Discussion}
\label{sec:discussion}

Personalized decision-making reflects a deep understanding of each customer's circumstances and precision interventions on the customer (client) for optimal objectives. This is challenging when dynamic, interactive and sequential decision-making processes are involved. In this work, personalized deep learning is proposed to learn and recommend next-best actions for each customer in the above context. We model the client-decision-maker interactions and their decision-making context related to client circumstances and behaviors and decision-maker actions and constraints. The proposed reinforced coupled recurrent network (CRN) provides a general neural multi-sequence interaction learning solution to formalize multi-party interactions with real-life evolving, long-term dependent states and behaviors of customers and intervention actions by decision-makers for automated personalized decision-making. The CRN incorporated with coupled recurrent units (CRU) effectively and efficiently models and recommends next-best actions for each client-oriented dynamic, personalized and sequential decision-making. CRU (1) reveals the complex long-term dependencies between client states, between decision actions, and between client responses and decision-maker interventions, and (2) involves and determines the client and decision-maker's historical information relevant to their responses and actions. In this way, we are able to model the complex multi-sequence interactions and coupling relationships between customer states and behaviors and decision-maker's actions and constraints for dynamic and personalized decision-making. This involves characterizing and coupling the roles, relationships and dynamics of clients and decision-makers in past, present and future decision-making processes. 

Our multi-sequence interaction learning method shows the potential of effectively modeling multi-aspect sequential, interactive and long-term dependencies, learning sequential historical information about a client's circumstances and sequential behavior responses to decision actions, and capturing the dynamic sequential interactions between client responses and decision actions. Our method, thus, goes beyond the usual way of assuming such decision-making processes as Markovian or convertible to Markovian \cite{bacchus1996rewarding,bacchus1997structured,thiebaux2006decision,ijcai_WangSC13,Lin2020}, which often only captures short-term dependencies in a single sequence and incurs a  high computational cost and a high rate of meaningless recommendations. Our method captures the above diverse multi-sequence-coupled and long-term dependencies while also controlling the computational cost. This explains why our method outperforms the wide-and-deep model, which utilizes the Markovian decision process.

We also show the potential of personalized decision-making by selecting actions for each client at each time point based on a deep representation of individual-level decision-making processes over time. CRN embeds the CRU-captured historical information and the current client state as a compact representation to learn decision rules, i.e., the dependencies between the current reward and historical states and actions. All these happen in a personalized and optimal manner, i.e., resulting in recommending next-best actions for each client at each time point per the then context. 

This study also goes beyond  non-Markovian decision process-based decision-making modeling \cite{gabaldon2011non,clarke2015human,friedrich2011spatio}, which models historical information by assuming a non-Markovian process but overlooks the sequential and multi-party interactions between stakeholders and between their behaviors. More research is required to further explore hierarchical, heterogeneous, time-varying and role-dependent couplings and interactions between multi-parties, between their behavior sequences, and between customer preferences and decision-making expectations. Multi-party interaction processes and dynamics also involve other challenges to be modeled, e.g., the imbalance in action distribution which may follow a Beta rather than a normal process in some applications, and hierarchical dependencies from attribute values to objects (e.g., clients), and the heterogeneities between customers. 

The recent advancements in RNNs with long short-term memory (LSTM) \cite{hochreiter1997long} and GRU has widely been applied to model the sequential decision processes \cite{bajor2017predicting}. They learn a representation for historical states and use the representation to inform decision-making to capture the dependencies between historical states and the current action. However, our method additionally captures the long-term interactions between actions and states and between actions and also incorporates the historical behaviors of clients into the current client states.

In addition, the recent work on sequential recommendation (such as next-item, next-basket and next-song \cite{WangS21,ji2015next,chou2016addressing} recommendation) and interactive recommendation \cite{WangC21-1} also involves contextual information. They typically apply neural networks and the attention mechanism \cite{bahdanau2014neural} to model contextual information related to the current object. Such methods cannot make next-best action recommendation since they do not involve decision processes, the impact evaluation of next actions, or dynamic environments, etc.

Further, interactive personalized decision-making needs to dynamically evaluate and optimize the reward of each decision action and recommend the next-best action in relation to a customer's current states, future rewards to actions, the customer's future responses, and decision objectives. We model the effect of each action on each client by considering a client's current context, past long-term behaviors, and decision feedback (effectiveness) on past actions measured by domain-driven rewards. This creates a way to involve domain knowledge, historical experience, and client and action-specific circumstances into a real-life complex decision-making process and interaction learning. 

Lastly, the pilot study on next-best actions for debt collection shows that modeling personalized, dynamic, sequential and interactive decision-making processes is often associated with diverse computational challenges. They include hierarchical imbalanced data distributions, multi-party interactions, and sequential, evolving, long-term and multi-sequence couplings and dependencies. Our neural interaction learning method paves a computational way to effectively and efficiently make personalized recommendations on next-best actions for a large number of clients in enterprise decision-making.

\section*{Conclusion}

Multi-party interactions involve multiple coupled sequences, e.g. of each party's states, behaviors and contexts. Personalized decision-making needs to not only model these coupled sequences and the couplings both within and between these sequences but also the couplings between parties, e.g., between a decision-maker and its clients. The automated learning of next-best actions to be taken on each customer at each time is essential for personalized and automated decision-making in any applications involving customer services and communications. Learning personalized next-best actions has to further model the multi-party interactions for each customer and his decision-maker and learn heterogeneous dynamic multi-sequence couplings. These issues go beyond classic decision theories, Markovian decision process theories, and sequential modeling and recommendation. User modeling, sequential modeling, behavior informatics, recommender systems and personalized decision-making should be integrated to address the challenges and complexities in learning automated decision-making with personalized next-best action recommendation and in dynamic, interactive and evolving personalized decision-making processes.

\bibliography{NBA-xiv}



\section*{Acknowledgements}


We acknowledge the funding support from the Australian Research Council Discovery grants (DP190101079 and FT190100734) and the series of projects contracted with the Australian Taxation Office.






\textbf{Competing interests} The authors declare that they have no competing financial interests.


\end{document}